\documentclass[conference]{IEEEtran}
\IEEEoverridecommandlockouts 
\usepackage{cite}
\usepackage{amsmath,amssymb,amsfonts}
\usepackage{graphicx}
\usepackage{textcomp}
\usepackage{xcolor}
\usepackage{algorithm}
\usepackage{algorithmic}
\usepackage[utf8]{inputenc}
\usepackage{booktabs}
\usepackage[T1]{fontenc}
\usepackage{amsmath,amsfonts,amsthm} 
\usepackage{hyperref}
\def\BibTeX{{\rm B\kern-.05em{\sc i\kern-.025em b}\kern-.08em
    T\kern-.1667em\lower.7ex\hbox{E}\kern-.125emX}}
\usepackage[normalem]{ulem} 

\makeatletter
\def\@maketitle{%
  \newpage
  \null
    \vskip 0.25in 
  \begin{center}%
    {\LARGE \bfseries \@title \par}%
    \vskip 2.0em
    {\large
      \begin{tabular}[t]{c}%
        \@author
      \end{tabular}\par}%
  \end{center}%
  \par
  \vskip 0em}
\makeatother

\begin{document}
\title{\fontsize{16pt}{18pt}\selectfont\bfseries Rao-Blackwellized POMDP Planning}
\author{\fontsize{11pt}{18pt}\selectfont Jiho Lee$^{1}$, Nisar Ahmed$^{1}$, Kyle Hollins Wray$^{2}$, Zachary Sunberg$^{1}$
\thanks{$^{1}$Ann And H.J. Smead Aerospace Engineering Sciences Department, University of Colorado Boulder, Boulder, CO, USA. {\tt\small Email: \{Jiho.Lee, Nisar.Ahmed, Zachary.Sunberg\}@colorado.edu}}
\thanks{$^{2}$Manning College of Information and Computer Sciences, University of Massachusetts Amherst, Amherst, MA, USA. {\tt\small Email: kwray@umass.edu}}}
\maketitle

\begin{abstract}
Partially Observable Markov Decision Processes (POMDPs) provide a structured framework for decision-making under uncertainty, but their application requires efficient belief updates. Sequential Importance Resampling Particle Filters (SIRPF), also known as Bootstrap Particle Filters, are commonly used as belief updaters in large approximate POMDP solvers, but they face challenges such as particle deprivation and high computational costs as the system's state dimension grows. To address these issues, this study introduces Rao-Blackwellized POMDP (RB-POMDP) approximate solvers and outlines generic methods to apply Rao-Blackwellization in both belief updates and online planning. We compare the performance of SIRPF and Rao-Blackwellized Particle Filters (RBPF) in a simulated localization problem where an agent navigates toward a target in a GPS-denied environment using POMCPOW and RB-POMCPOW planners. Our results not only confirm that RBPFs maintain efficient belief approximations over time with fewer particles, but, more surprisingly, RBPFs combined with quadrature-based integration improve planning quality significantly compared to SIRPF-based planning under the same computational limits.
\end{abstract}

\section{Introduction}
Partially Observable Markov Decision Processes (POMDPs) are a powerful mathematical framework for modeling decision-making under uncertainty where an agent operates in an environment with incomplete or noisy information \cite{c1}. POMDPs have been widely applied to various domains such as aircraft collision avoidance, automated driving, and search-and-rescue with drones \cite{c2}\cite{c17}\cite{c18}\cite{c19}. However, one of the key challenges in implementing POMDPs is the need for efficient belief updates that maintains a reliable probability distribution across the system's possible states.

Particle filters are effective for modeling the agent's beliefs when the state space is large or continuous \cite{c4}\cite{c5}\cite{c16}. However, particle filters often suffer from particle deprivation where the diversity of particles diminishes over time. To address this issue, Sequential Importance Resampling Particle Filtering (SIRPF), also known as Bootstrap Particle Filtering, is commonly employed. SIRPF mitigates particle deprivation by resampling particles based on their weights, discarding low-weight and duplicating high-weight particles. Additionally, various strategies, such as adaptive particle injection and rejection, have been used to mitigate the deprivation issue \cite{c1}. While SIRPF is effective in many scenarios, it still remains sensitive to outliers and unlikely observations, leading to a low Effective Sample Size (ESS) where only a small fraction of particles contribute to the state estimate. Moreover, as the system's effective dimension grows, a substantial increase in the number of particles may be required to maintain performance, resulting in high computational costs (e.g. \cite{c3}).

\begin{figure}[!t]
\centerline{\includegraphics[width=1.0\columnwidth]{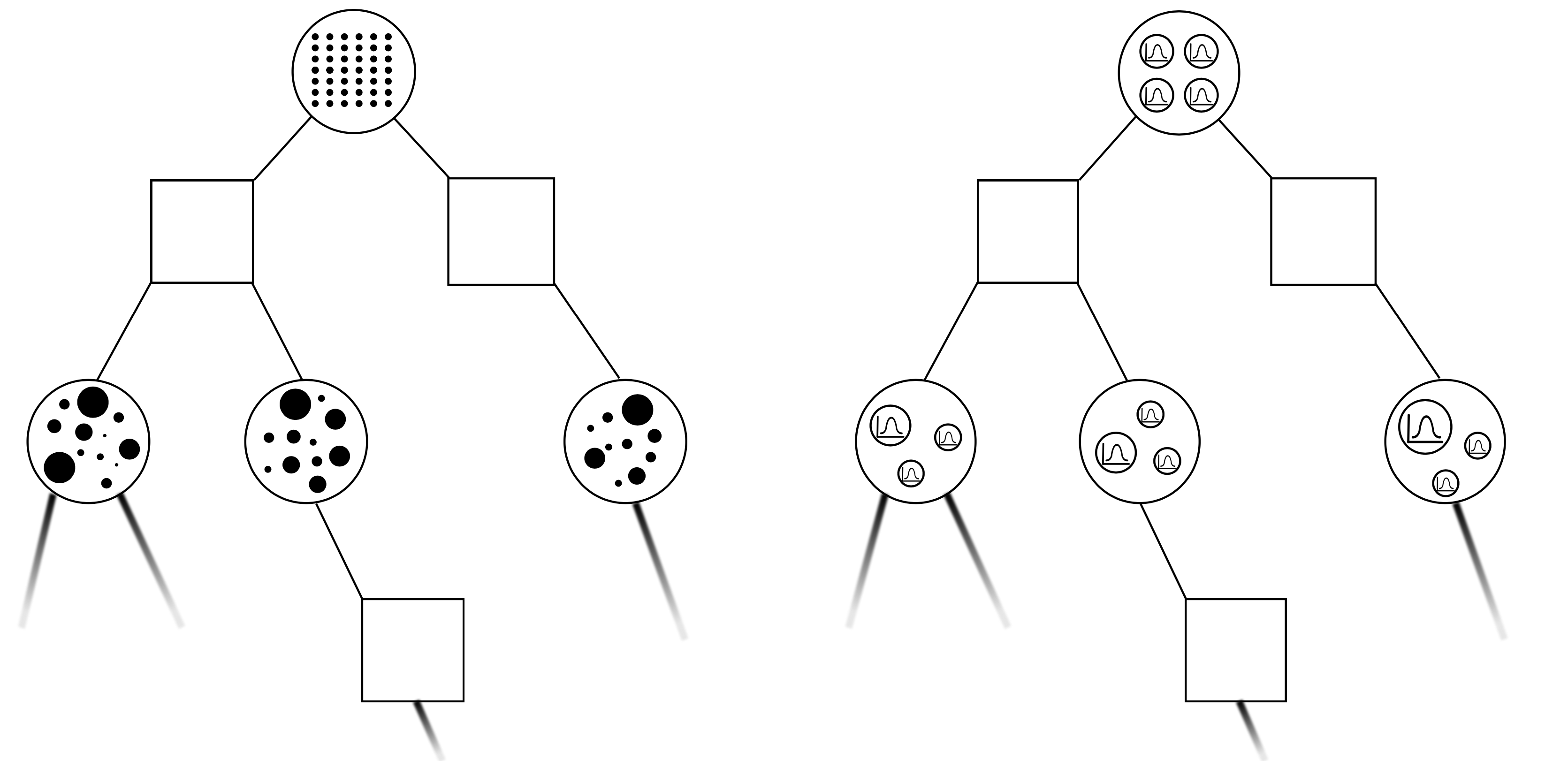}}
\caption{POMCPOW (left) and RB-POMCPOW (right) Tree Structure Comparison. Each square and larger circle represents an action node and an observation node, respectively. In the POMCPOW tree, particles are shown as black dots while each particle in the RB-POMCPOW tree is associated with a Gaussian distribution. Due to the nature of POMCPOW, these particles form a weighted mixture of beliefs.}
\vspace{-0.4cm}
\label{fig:rbpf_tree_comparison}
\end{figure}

Rao-Blackwellized Particle Filtering (RBPF) offer a promising solution to address some of these limitations of the SIRPF. RBPF combines particle filtering with analytical methods, such as Kalman filters or Hidden Markov Model (HMM) filters \cite{c24}. By analytically solving for belief updates within the "tractable substructure" of the model, RBPF can reduce the dimensionality that particles need to explore and lower the overall number of particles required to approximate complex belief distributions\cite{c14}. Thus, by utilizing this hybrid approach, RBPF can reduce computational costs without compromising accuracy. 

While RBPF has been extensively studied in domains such as Simultaneous Localization and Mapping (SLAM), target tracking, and entropy-driven exploration \cite{c14}\cite{c20}\cite{c21}\cite{c22}, its application within the full context of POMDPs is very limited \cite{c27}. Within POMDPs, agents not only need to estimate their current state but also must strategically plan actions that consider both immediate and long-term rewards. In this study, we make three key contributions. First, we leverage RBPF as the belief updater within the approximate POMDP solution framework which we refer to as Rao-Blackwellized POMDP (RB-POMDP) that enables the use of analytical filters for conditionally dependent tractable states. Second, we introduce a new planning algorithm, Rao-Blackwellized partially observable Monte Carlo planning with observation widening (RB-POMCPOW). This new planner is designed to handle continuous state and observation spaces as well as the uncertainty in the analytically marginalized distributions of Rao-Blackwellized particles, as depicted in Figure \ref{fig:rbpf_tree_comparison}. Also, the planner reduces the variance in value estimates through quadrature integration methods and has the flexibility to allow practitioners to select appropriate integration techniques based on the specific distributions of those marginalized states. Finally, we present a comparative analysis of SIRPFs and RBPFs in a localization problem where an agent navigates toward a target in a GPS-denied environment using POMCPOW and RB-POMCPOW as the respective planners. Our results demonstrate that the RBPF with 100 particles achieves better ESS and cumulative rewards than the SIRPF with 1000 particles while also being approximately seven times faster in planning when using effective quadrature points. Furthermore, by increasing the number of quadrature points and thereby improving the accuracy of the integration methods, the RBPF was shown to outperform the SIRPF under the same computational time. In conclusion, this study underscores the RBPF's computational advantages in both planning and belief updates and provides an efficient framework for more complex decision-making environments.

\section{Background}

\subsection{Partially Observable Markov Decision Process (POMDP)}
In Markov Decision Processes (MDPs), agents operate with complete knowledge of the current state of the environment. At any given time, the agent fully understands where it is or what the situation is despite uncertainties in future state transitions. In POMDPs, however, agents do not have perfect knowledge of the current state. They must make decisions under uncertainty about the current as well as future states; thus, POMDPs are more suitable for real-world robotics scenarios where perfect state information is rarely available \cite{c26}.

A POMDP is formally defined by the tuple $(\mathcal{S}, \mathcal{A}, \mathcal{T}, \mathcal{O}, \mathcal{R}, \mathcal{Z}, \gamma)$ where $\mathcal{S}$ is the set of possible states, $\mathcal{A}$ is the set of actions, $\mathcal{O}$ is the set of possible observations, $\mathcal{T}$ is the state transition function, $\mathcal{R}$ is the reward function, $\mathcal{Z}$ is the observation function, and \( \gamma \) is the discount factor that balances between immediate and future rewards. Because the agent lacks full visibility of the state, it maintains a probabilistic \emph{belief} over the current possible states of the environment. This belief is updated using the following Bayesian equation:
\begin{equation}
\nonumber
b'(s') 	\propto P(o | s', a) \sum_{s \in S} P(s' | s, a) b(s),
\end{equation}
where $b'(s')$ is the posterior belief of being in state $s'$ after taking action $a$ and receiving observation $o$, and $b(s)$ is the prior belief of being in state $s$.

Solving POMDPs involves finding optimal policies that specify the best action to take based on the current belief. The value function, which guides the selection of the best action, can be expressed as:
\[
V^*(b) = \max_{a \in A} \left[  \mathcal{R}(b, a) + \gamma \sum_{o \in O} P(o | b, a) V^*(b') \right],
\]
where $V^*(b)$ is the optimal value of the belief state $b$, $ \mathcal{R}(b, a)$ is the expected immediate reward for taking action $a$, $P(o | b, a)$ is the probability of receiving observation $o$ after taking action $a$, and $b'$ is the updated belief. Hence, by continuously updating its belief, the agent can make more informed decisions in uncertain environments. 

\subsection{Rao-Blackwellized Particle Filter (RBPF)}
Rao-Blackwellized Particle Filtering (RBPF) is an advanced variant of particle filtering that leverages the Rao-Blackwell theorem to enhance computational efficiency and estimation accuracy. This theorem is based on sufficient statistics to reduce the variance of estimators \cite{c13} as represented by the following inequality:
\begin{equation}
\nonumber
    \text{var}[\delta(s)] \geq \text{var}[\delta(s |\Theta)].
\end{equation}
Here, $\delta$ is any kind of estimator of $s$, and $\Theta$ represents the sufficient statistics for $s$. Essentially, conditioning on $\Theta$ retains all necessary information about $s$, which can only reduce the variance of the estimator $\delta$. Applying this principle to Monte Carlo-based estimators, RBPF can use fewer particles by incorporating sufficient statistics to capture some state parameters analytically.

\section{Technical Approach}
Several adjustments are necessary to implement analytical filters within sampling-based approximate POMDP solvers, an approach we dub the RB-POMDP framework. In an RBPF, each particle is not just a discrete state sample but is also associated with a conditional analytical distribution. This requires modifications in both the belief update process and the planning algorithm. This section describes our novel approach to address both these issues.

\subsection{Rao-Blackwell factorization of POMDPs}
The main idea in RPBF is to factorize the state space into two components: \emph{tractable} and \emph{non-tractable} states. More formally, the joint posterior distribution of these tractable and non-tractable components can be factored into the following equation using the chain rule: 
\begin{equation}
\label{eq: factorization}
    p(s_{k+1}^\alpha, s_{k+1}^\pi \mid o_{1:k}) = p(s_{k+1}^\alpha \mid s_{k+1}^\pi, o_{1:k}) p(s_{k+1}^\pi \mid o_{1:k})
\end{equation}
The RBPF uses this relationship to update the marginalized distribution of the tractable state $s_{k+1}^\alpha$ analytically using methods like Kalman filters, while the non-tractable state $s_{k+1}^\pi$ is updated through particle filtering. Consequently, each particle in RBPF is associated with a conditional analytical distribution, which in many practical cases can be summarized by sufficient statistics.

Equation \ref{eq: factorization} implies that if the state space can be correctly decomposed into non-tractable and tractable components, then we can leverage RBPF within POMDP models. This approach allows RBPFs to focus the particle filter sampling only on the non-tractable component, reducing the overall number of particles.

\subsection{RBPF Belief Updates}
The most common choice for the importance distribution in the sequential importance sampling step is the transition function $p(s'|s,a)$ which simplifies the importance weight to be equal to the observation likelihood \cite{c15}. For RBPFs, the weights are updated based on this observation likelihood (denoted by $\Lambda$ in line 3 of Algorithm \ref{alg:belief_update}) that is calculated using the analytical method applied to the tractable components \cite{c23}. When using a Kalman filter, for example, the update involves the innovation covariance matrix. This approach differs from the standard SIRPF where the weights are updated solely based on the likelihood of the observation and does not incorporate an analytical step. Hence, the weight update in RBPFs leverages sufficient statistics ($\Theta$), resulting in more efficient updates and lower variance, as guaranteed by the Rao-Blackwell theorem, compared to SIRPFs. The RBPF belief update is detailed in Algorithm \ref{alg:belief_update}.

\subsection{RBPF in Sampling-Based Online Planners}
Partially observable Monte Carlo planning (POMCP), a widely used online planner, and its variant POMCPOW rely on Monte Carlo sampling via the particle filter to estimate the value function \cite{c4}\cite{c5}. As shown in Fig. \ref{fig:rbpf_tree_comparison} and Alg.~\ref{alg:RB-POMCPOW}, these planners build a local policy from the agent's current belief state by running N tree search simulations to a depth $D$. The value at an observation-action history $\tau$ steps into the future, $h$ is estimated based on the rewards for the set of simulations consistent with that history, $I(h)$, leading to the following value approximation:
\begin{equation*}
V(h) \approx \frac{1}{|I(h)|} \sum_{i\in I(h)} \sum_{d=\tau}^{D} \gamma^{d-\tau}  \mathcal{R}(s_{i,d}, a_{i,d}).
\end{equation*}
Monte Carlo sampling, however, is known for its slow convergence rate of \(\mathcal{O}(1/\sqrt{N})\). As a result, this approach can be computationally heavy and can require a large number of samples to achieve accurate estimates. By leveraging RBPF, we can compute expectations over the tractable components using deterministic numerical integration methods (e.g. quadrature methods) such that the overall number of Monte Carlo samples (i.e. tree iterations) can be reduced, as detailed below. 

\begin{algorithm}[!t]
\caption{Belief Updates for Rao-Blackwellized Particle Filter (RBPF) adapted from SIRPF algorithm in \cite{c13}}
\label{alg:belief_update}
\begin{algorithmic}[1]
\REQUIRE $(s_k^{\pi,i}, w_k^{i}, \Theta_k^{\alpha|\pi,i})_{i=1}^{N_s}$, $o_{k+1}, a_{k+1} $
\ENSURE$(s_{k+1}^{\pi,i}, w_{k+1}^{i}, \Theta_{k+1}^{\alpha|\pi,i})_{i=1}^{N_s}$
\FOR{$i = 1$ \TO $N_s$}
    \STATE \text{Draw} $s_{k+1}^{\pi,i} \sim p(s_{k+1}^{\pi,i} | s_k^{\pi,i}, a_{k+1})$
    \STATE $\Theta_{k+1}^{\alpha|\pi,i}, \Lambda_{k+1}^{\alpha|\pi,i} \gets \text{AnalytUpd}\left(\Theta_k^{\alpha|\pi,i}, s_{k+1}^{\pi,i}, o_{k+1},a_{k+1}\right)$
    \STATE $\tilde{w}_{k+1}^{i} \gets \text{Reweight}\left(\Theta_{k+1}^{\alpha|\pi,i}, \Lambda_{k+1}^{\alpha|\pi,i}, s_{k+1}^{\pi,i}, o_{k+1}\right)$
\ENDFOR

\STATE $\left\{ w_{k+1}^{i} \right\}_{i=1}^{N_s} = \text{Normalize}\left( \left\{ \tilde{w}_{k+1}^{i} \right\}_{i=1}^{N_s} \right)$
\STATE \text{Compute} $N_{\text{ess}} = \frac{1}{\sum_{s=1}^{N_s} (w_{k+1}^{i})^2}$
\IF {$N_{\text{ess}} < \tau$}
    \STATE $\left\{ s_{k+1}^{\pi,i}, w_{k+1}^i \right\}_{i=1}^{N_s} \gets \text{Resample} \left( \left\{ s_{k+1}^{\pi,i}, w_{k+1}^i \right\}_{i=1}^{N_s} \right)$
\ENDIF
\end{algorithmic}
\end{algorithm}

\subsection{Rao-Blackwellized POMCPOW (RB-POMCPOW)}
\label{section:RB-POMCPOW}

The standard logic of POMCP and POMCPOW solvers can be adapted to effectively utilize the analytical distributions that each Rao-Blackwellized particle carries as depicted in Figure \ref{fig:rbpf_tree_comparison}. To begin, consider the immediate reward for the states consistent with $h$ approximated as follows:
\begin{equation*}
      \mathcal{R}(h) \approx 
     \frac{1}{|I(h)|} \sum_{i \in I(h)} \underset{s_{i}^{\alpha|\pi}}{\mathbb{E}} \left[ \mathcal{R}(s_{i}^\pi, s_{i}^{\alpha|\pi},a_{i}) \right].   
\end{equation*}
Here, the expectation is taken over $s_{i}^{\alpha|\pi}$ to account for the uncertainty in the analytical distributions. In other words, we need to employ appropriate integration methods that allow us to compute the value function in a manner that incorporates different possible realizations of the tractable state components. Next, we extend the approximation over a multi-step planning horizon of depth $D$ where the agent accumulates discounted rewards from future actions $a_{i,d}$: 
\begin{equation*}
    V(h) \approx \frac{1}{|I(h)|} \sum_{i\in I(h)} \sum_{d=\tau}^{D} \gamma^{d-\tau}  \underset{s_{i,d}^{\alpha|\pi}}{\mathbb{E}} \left[  \mathcal{R}(s_{i,d}^{\pi}, s_{i,d}^{\alpha|\pi}, a_{i,d}) \right] . \\ 
\end{equation*}
While straightforward Monte Carlo sampling can also be used here to compute these expectations, deterministic quadrature techniques (e.g., tensor product quadrature) offer the advantage of faster convergence and higher accuracy, particularly for lower dimensional integrals. Each expectation is approximated using a set of $M$ quadrature points $s_{i,d,k}^{\alpha|\pi}$ and corresponding weights $w_{i,d,k}$, enabling us to replace the expectation over the tractable components with a weighted sum: 
\begin{equation*}
    V(h) \approx \frac{1}{|I(h)|} \sum_{i\in I(h)} \sum_{d=\tau}^{D} \gamma^{d-\tau} \sum_{k=1}^{M} w_{i,d,k} \,  \mathcal{R}(s_{i,d}^{\pi}, s_{i,d,k}^{\alpha|\pi}, a_{i,d}).
\end{equation*}
When using quadrature techniques, the choice of method depends on the distribution used for the analytically marginalized states. For example, we can employ Gaussian-Hermite quadrature to interpolate and perform the necessary integration when using Gaussian beliefs for the linear states\cite{c8}. If different distributions were considered, other methods from the Askey family of orthogonal polynomials could be utilized \cite{c6}\cite{c7}. To mitigate the curse of dimensionality, we can implement a Smolyak sparse grid, which efficiently reduces the number of quadrature points required while maintaining accuracy \cite{c9}\cite{c10}. The Smolyak formula for constructing the sparse grid is: 
\[
\mathcal{A}(q,d) = \sum_{q-d+1 \leq |\mathbf{i}| \leq q} (-1)^{q - |\mathbf{i}|} \binom{d-1}{q - |\mathbf{i}|} \bigotimes_{j=1}^{d} \mathcal{Q}_{i_j},
\] where $\mathcal{A}(q,d)$ is the set of Smolyak quadrature points and their corresponding weights for a given sparse grid level $q$ and dimensionality $d$, $i$ is a multi-index with $i_j$ representing the level of the univariate quadrature rules, and $ \bigotimes_{j=1}^{d} \mathcal{Q}_{i_j}$ denotes the tensor product of the univariate quadrature rules $\mathcal{Q}_{i_j}$ at levels $i_j$ for each dimension $j$.

A higher level of the sparse grid results in greater accuracy but requires more computational time. By choosing different levels of sparse grid, we can fine-tune the balance between computational cost and accuracy in the planner. Overall, this approach reduces the reliance on Monte Carlo sampling for all states by capturing the marginalized states using deterministic quadrature techniques and ultimately reduces the total number of tree iterations needed in POMCP and POMCPOW.

The RB-POMCPOW pseudocode, adapted from \cite{c5}, is presented in Algorithm \ref{alg:RB-POMCPOW} with the modified code written in blue. Quadrature techniques can be used in lines $17$, $27$, and $28$. Note that the sparse grid level or the integration methods can be adjusted independently for the $\textsc{Rollout}$ and $\textsc{Simulate}$ functions to compute the expectations. While this study primarily focuses on the RB-POMCPOW algorithm, the underlying principles for handling uncertainty in the analytical distributions of tractable states remain the same for POMCP (refer to Appendix \ref{appendix} for the pseudocode).

\begin{algorithm*}[ht]
\caption{RB-POMCPOW adapted from \cite{c5}}
\label{alg:RB-POMCPOW}
{\small
\begin{algorithmic}[1]
\begin{minipage}[t]{0.48\textwidth}
\STATE \textbf{procedure} \textsc{Search}($h$)
    \STATE \quad \textbf{for} $i \leftarrow 1 \textbf{ to } N$ \textbf{do}
        \STATE \quad \quad \textcolor{blue}{\textsc{Simulate}($p \sim  b(h), h, d_{max}$)}
    \STATE \quad \textbf{return} $\arg\max_a Q(ha)$;
\STATE
\STATE \textbf{procedure} \textsc{ActionProgWiden}$(h)$
    \STATE \quad \textbf{if} $|\mathcal{C}(h)| \leq k_a N(h)^{\alpha_a}$ \textbf{then}
    \STATE \quad \quad $a \leftarrow \textsc{NextAction}(h)$
    \STATE \quad \quad $\mathcal{C}(h) \leftarrow \mathcal{C}(h) \cup \{a\}$
    \STATE \quad \textbf{return} ${\arg\max\limits}_{a \in \mathcal{C}(h)} \left[ Q(ha) + c\sqrt{\frac{\log N(h)}{N(ha)}} \right]$
\STATE
\STATE \textbf{procedure} \textsc{Rollout}($p, h, d$)
    \STATE \quad \textbf{if} $\gamma^{\text{depth}} < \epsilon$ \textbf{then}
        \STATE \quad \quad \textbf{return} 0
    \STATE \quad \textbf{end if}
    \STATE \quad $a \sim \pi_{\text{rollout}}(h, \cdot)$
    \STATE \quad \textcolor{blue}{$(\hat{s'}, \hat{o}, \hat{r}) \leftarrow \mathbb{E}[G(p, a)]$} \hfill \textcolor{black}{$\triangleright$ Quadrature Techniques} \phantom{xx}
    \STATE \quad \textcolor{blue}{$p' \leftarrow$ \textsc{AnalyticalUpdate}$(p, \hat{s'}, \hat{o}, a)$} 
    \STATE \quad \textcolor{blue}{\textbf{return} $\hat{r} + \gamma \cdot \textsc{Rollout}(p', ha\hat{o}, d - 1)$}
\STATE \textbf{end procedure}
\STATE 
\STATE \textbf{procedure} \textsc{Simulate}($p, h, d$)
    \STATE \quad \textbf{if} $d = 0$ \textbf{then}
        \STATE \quad \quad \textbf{return} 0
    \STATE \quad \textbf{end if}
\end{minipage}
\hfill
\begin{minipage}[t]{0.48\textwidth}
    \STATE \quad $a \leftarrow \textsc{ActionProgWiden}(h)$
    \STATE \quad \textcolor{blue}{$\hat{s} \leftarrow \mathbb{E}[p]$} \hfill \textcolor{black}{$\triangleright$ Quadrature Techniques}
   \STATE \quad \textcolor{blue}{$(\hat{s'}, \hat{o}, \hat{r}) \leftarrow \mathbb{E}[G(p, a)]$} \hfill \textcolor{black}{$\triangleright$ Quadrature Techniques}
    \STATE \quad \textbf{if} $|C(ha)| \le k_o N(ha)^{\alpha_o}$ \textbf{then}
        \STATE \quad \quad \textcolor{blue}{$o \leftarrow \hat{o}$}
        \STATE \quad \quad $M(hao) \leftarrow M(hao) + 1$
    \STATE \quad \textbf{else}
        \STATE \quad \quad $o \leftarrow \text{select } o \in C(ha) \text{ w.p. } \frac{M(hao)}{\sum_{o} M(hao)}$
    \STATE \quad \textbf{end if}
    \STATE \quad  \textcolor{blue}{\textbf{append} $(\hat{s'}, \hat{r})$ \textbf{to} $B(hao)$}
    \STATE \quad \textcolor{blue}{\textbf{append} $\mathcal{Z}(o \mid \hat{s}, a, \hat{s'})$ \textbf{to} $W(hao)$}
    \STATE \quad \textbf{if} $o \notin C(ha)$ \textbf{then} \quad \textit{ $\triangleright$ new node}
        \STATE \quad \quad $C(ha) \leftarrow C(ha) \cup \{o\}$
        \STATE \quad \quad \textcolor{blue}{$p' \leftarrow$\textsc{AnalyticalUpdate}$(p, \hat{s'}, \hat{o}, a)$}
\STATE \quad \quad \textcolor{blue}{$total \leftarrow \hat{r} + \gamma$ * \textsc{Rollout}($p'$, $hao$, $d - 1$)}
    \STATE \quad \textbf{else}
        \STATE \quad \quad \textcolor{blue}{$(s',r) \leftarrow \text{select } B(hao)[i] \text{ w.p. } \frac{W(hao)[i]}{\sum_{j=1}^{m} W(hao)[j]}$}
         \STATE \quad \quad \textcolor{blue}{$p' \leftarrow$\textsc{AnalyticalUpdate}$(p, s', o, a)$}
        \STATE \quad \quad \textcolor{blue}{$total \leftarrow r + \gamma \textsc{Simulate}(p', hao, d - 1)$}
    \STATE \quad \textbf{end if}
    \STATE \quad $N(h) \leftarrow N(h) + 1$
    \STATE \quad $N(ha) \leftarrow N(ha) + 1$
    \STATE \quad $Q(ha) \leftarrow Q(ha) + \frac{total - Q(ha)}{N(ha)}$
    \STATE \quad \textbf{return} $total$
\STATE \textbf{end procedure}
\end{minipage}
\end{algorithmic}}
\end{algorithm*}
\section{Experiments}
\vspace{-0.08cm}
This section focuses on a specific localization problem, but the approach presented here can be applied to a broader class of planning problems. We used the POMDPs.jl \cite{c25} in implementing the methods presented in this work.
\subsection{Localization Problem}
The goal of our planning module is to devise a navigational strategy for an agent in a 2-dimensional environment to reach the target location at the origin while avoiding the obstacle. Additionally, the agent operates in a GPS-denied environment, making it a localization problem where the agent predicts its own position based on noisy observations from static landmarks. This section details the elements of the POMDP model $ \mathcal{P}=(\mathcal{S}, \mathcal{A}, \mathcal{T}, \mathcal{O}, \mathcal{R}, \mathcal{Z})$ that we utilize for our study.

\textbf{State Space $\mathcal{S}$:} The state space consists of continuous variables representing the agent's position and orientation. Specifically, the state vector is defined as \( s = \{ \xi, \eta, \theta\} \), where \( \xi\) and \( \eta\) are the agent's coordinates and \( \theta \) is its heading.

\textbf{Actions \( \mathcal{A} \):} The action set available to the agent includes different movement modes, represented as tuples \( (s_t, w_t) \) where \( s_t \) is the speed and  \( w_t \) is the turning rate. The agent can move forward, backward, or turn left and right by adjusting these parameters.

\textbf{Observations and Observation Model $\mathcal{O}, \mathcal{Z}$:} The observation space consists of noisy sensor measurements of known landmarks. The observation vector is defined as $\mathbf{o} = \{(\rho_j, \phi_j)\}_{j=1}^J$
where \(\rho_j\) and \(\phi_j\) are relative range and bearing to each landmark $j$, respectively, and \( J \) is the total number of landmarks the agent uses for localization. The observation model includes these noisy relative range \(\rho_j\) and bearing \(\phi_j\) measurements which are modeled as:
    \begin{align*}
      \rho_j(k) &= \sqrt{(\xi_{j} - \xi(k))^2 + (\eta_{j} - \eta(k))^2} + v_{r}(k), \\
      \phi_j(k) &= \operatorname{atan2}(\eta_{j} - \eta(k), \xi_{j} - \xi(k)) - \theta(k) + v_{b}(k),
    \end{align*}
where \( v_r(k) \sim \mathcal{N}(0, 0.25) \) and \( v_b(k) \sim \mathcal{N}(0, 0.01) \) represent the additive white Gaussian noise (AWGN) components in the range and bearing measurements, respectively.

\textbf{Transition Model $\mathcal{T}$:}  The dynamics of the agent are described by the following differential equations:
\begin{align*}
\dot{\xi} &= s_t \cdot \tilde{w}_s \cdot \cos(\theta), \\
\dot{\eta} &= s_t \cdot \tilde{w}_s \cdot \sin(\theta), \\
\dot{\theta} &= \omega_t \cdot \tilde{w}_{\omega}.
\end{align*}
Here, \( \tilde{w}_s \) and \( \tilde{w}_{\omega} \) represent the noise components, distributed as \(N_w([1; 1]^T, 0.1 \cdot I_{2 \times 2}) \), affecting the speed and turning rate, respectively. The variables $s_t$ and $w_t$ are defined by the chosen action. Note that $\dot{\xi}$ and $\dot{\eta}$ are dependent on $\theta$.

\textbf{Reward Model \( \mathcal{R} \):} The reward function is defined as quadratic to penalize deviations from the desired state (the origin) and the control input exerted by the agent, while also providing a terminal reward for successfully reaching the goal. The overall reward function is defined as:
\[
\mathcal{R}(s, a) = - \left( s^\top \Psi_s s + a^\top \Phi_a a \right) + \mathcal{R}_{\text{terminal}},
\]
where \( \Psi_s \) and \( \Phi_a \) are weighting matrices.

\begin{figure}[!t]
\centerline{\includegraphics[width=0.8 \columnwidth]{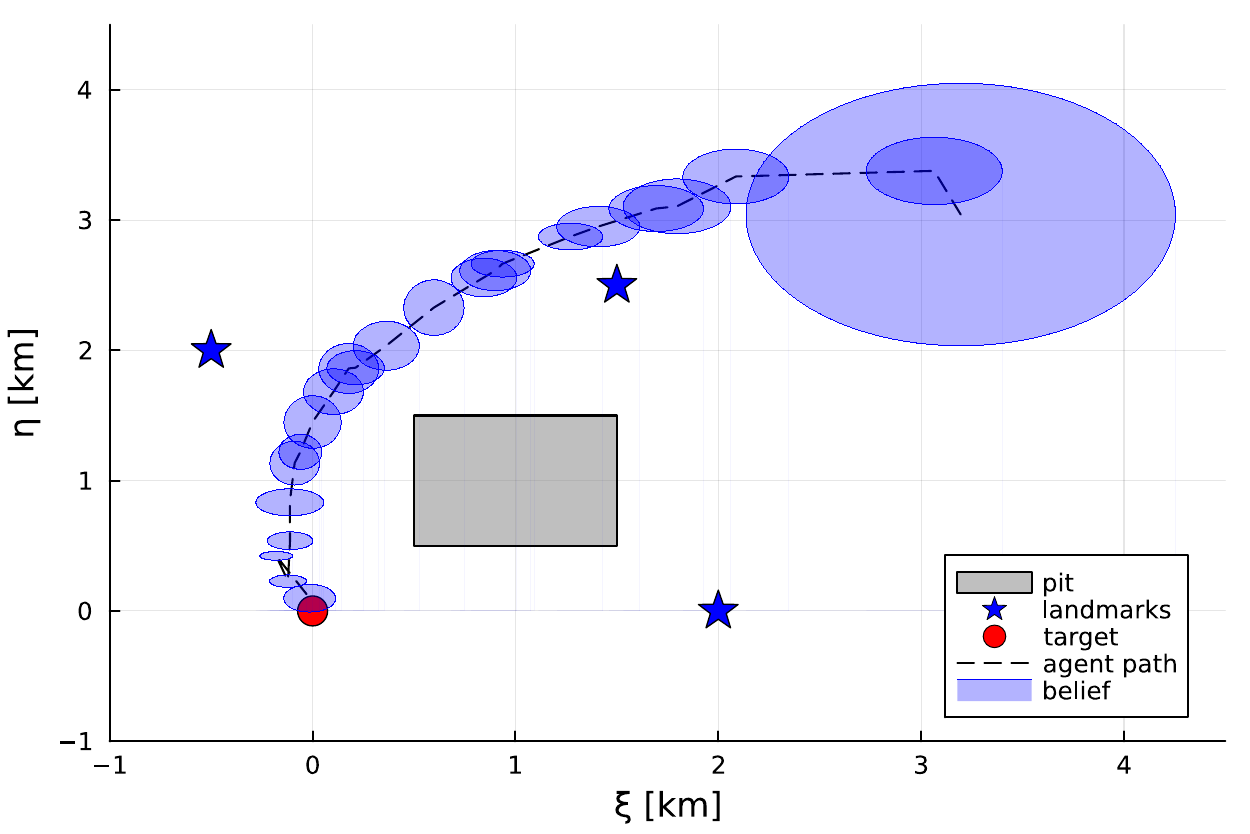}}
\caption{POMDP Localization Problem where the agent navigates toward the target. The agent receives noisy measurements from the static landmarks.}
\vspace{-0.4cm}
\label{fig:POMDP_localization}
\end{figure}
For our problem, we selected POMCPOW over POMCP for online planning due to its ability to handle continuous observation spaces through Observation Widening (OW)\cite{c5}. Also, the Unscented Kalman Filter (UKF) \cite{c12} is used to derive the Gaussian belief $\mathcal{N}(\mu, \Sigma)$ for $\xi$ and $\eta$ when given a predicted $\theta$. We conducted 100 simulations to analyze and compare the performance of SIRPF and RBPF in our localization problem by using POMCPOW and RB-POMCPOW as the respective planners. A representative simulation is depicted in Figure \ref{fig:POMDP_localization}.

\subsection{Filter Performance Testing}
\label{sec:filter_performance_testing}
\begin{figure}[t!]
\centerline{\includegraphics[width=\columnwidth]{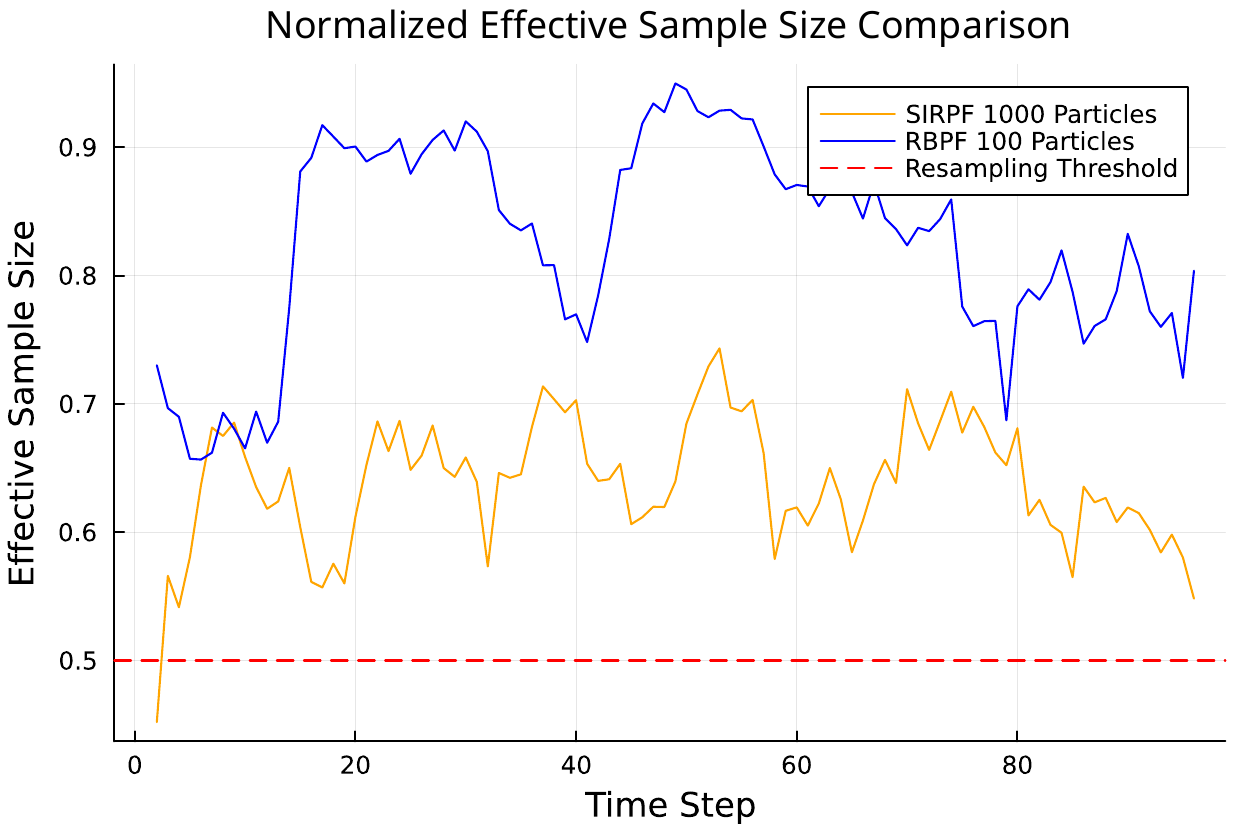}}
\vspace{-0.2cm}
\caption{$N_{ESS}$ comparison between RBPF with 100 particles and SIRPF with 1000 particles, averaged over 100 simulations and smoothed using a moving average to reduce noise and visualize trends. The red dashed line represents the threshold that indicates our minimum desired efficiency for resampling.}
\vspace{-0.4cm}
\label{fig:ess_comparison}
\end{figure}

RBPF enables analytical consistency testing that provides quantitative metrics to assess the filter's accuracy and consistency for the tractable states. This is an advantage over SIRPF which relies solely on particles for all state components. To ensure the filter's consistency, we conducted Normalized Estimation Error Squared and Normalized Innovation Squared tests \cite{c28} and fine-tuned the UKFs within the RBPF. Following this, we evaluated the efficiency of the particles in each filter using the Effective Sample Size (ESS) as illustrated in Figure \ref{fig:ess_comparison}. ESS quantifies the number of effective particles in the particle set based on the variance of their weights. Therefore, a higher ESS indicates more particles contribute meaningfully to the belief update whereas a low ESS suggests particle degeneracy where most particles have negligible weights. Normalized ESS is calculated as \cite{c13}:
\begin{equation}
\nonumber
N_{ESS} = \frac{1}{\sum_{i=1}^N w_i^2},
\end{equation}
where $w_i$'s are the normalized particle weights.

The $N_{ESS}$ for both SIRPF and RBPF was constantly above the 50 percent threshold that we set for our resampling requirement. However, the RBPF with 100 particles generally maintained a higher $N_{ESS}$ than the SIRPF with 1000 particles, indicating that RBPF produces more informed weight updates. This results in lower variance in importance weights and improves estimation efficiency for $\theta$. 

\subsection{Computational Costs and Cumulative Rewards}
We focus on two key aspects of the computational costs: the time required for belief updates and the time needed for the planner's action selection. From Table \ref{tab:belief_update}, it is evident that RBPF is slower than SIRPF for belief updates when using the same number of particles. This difference is attributed to the increased computational complexity involved in using UKFs and analytically updating the Gaussian belief associated with each particle. Expectedly, the time taken for belief updates increases linearly with the number of particles. However, it was observed that both RBPF and SIRPF achieved stable state estimation accuracy with low mean squared error when using 100 and 1000 particles, respectively, while maintaining high $N_{ESS}$. This informed our decision to use 100 particles for RBPF and 1000 particles for SIRPF. Consequently, this choice provided computational benefits during belief updates, with RBPF taking only $9$ ms per step compared to $32$ ms for SIRPF.  


Next, we analyzed the impact of value function accuracy on cumulative rewards. In the RB-POMCPOW algorithm, we varied the sparse grid level (denoted by $q$) in the \textsc{Simulate} function while simply using the mean $(\mu)$ of the Gaussian distribution for the tractable states in the \textsc{Rollout} function to reduce computational complexity. Furthermore, since the tractable components are now captured using quadrature points during value function computation, we reduced the number of tree iterations to 50 for RB-POMCPOW. 

\begin{figure}[!t]
\centerline{\includegraphics[width=\columnwidth]{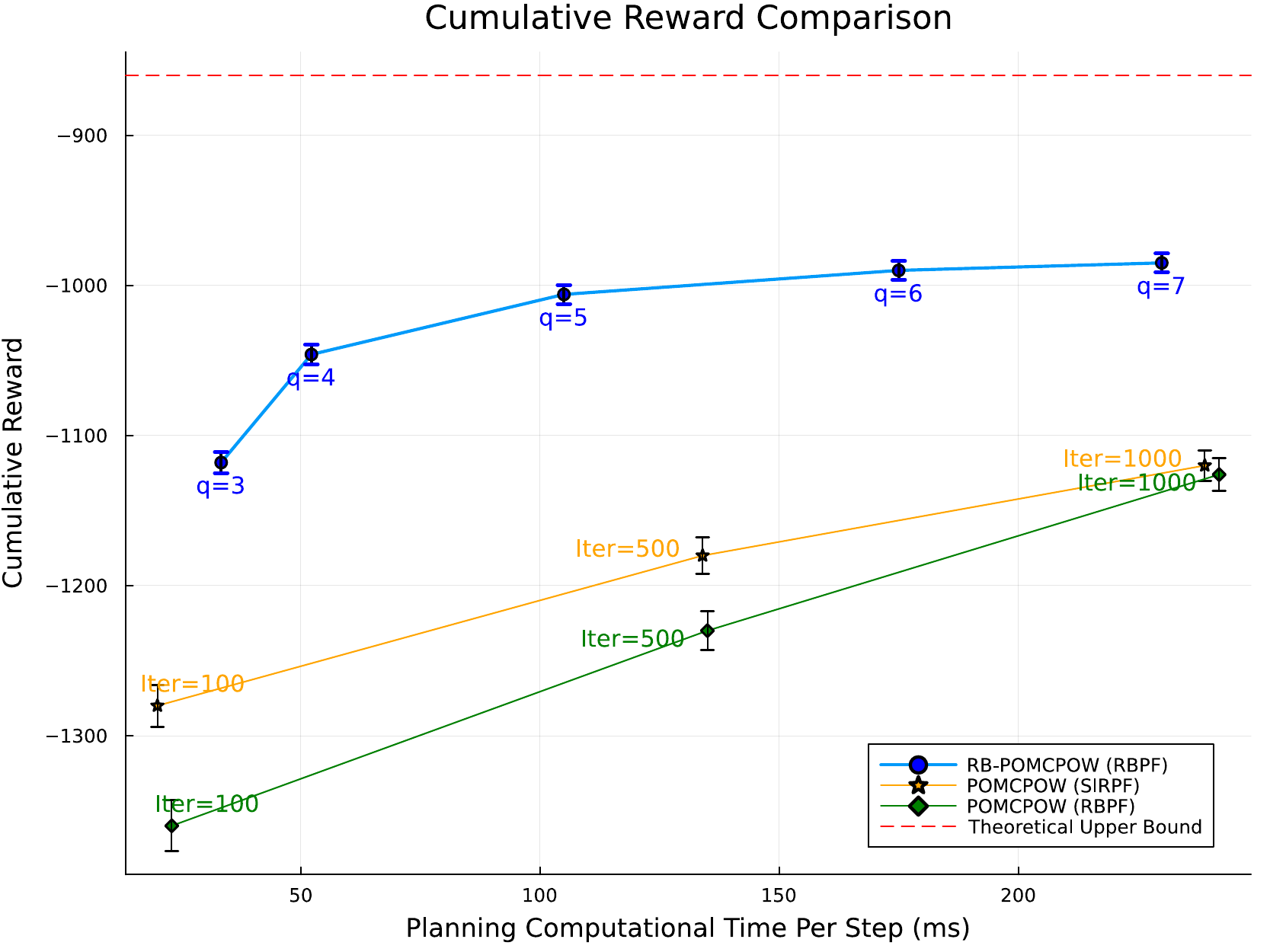}}
\caption{Cumulative reward comparison between RBPF and SIRPF in RB-POMCPOW and POMCPOW. The graph shows cumulative rewards for different sparse grid levels $(q)$ in RB-POMCPOW (with a fixed 50 tree iterations) and varying tree iterations in POMCPOW using SIRPF and RBPF. A theoretical upper bound on the cumulative reward is computed by assuming an ideal scenario where the agent aligns its heading perfectly and moves directly to the target without encountering obstacles.}
\vspace{-0.5cm}
\label{fig:computational_cost}
\end{figure}

As shown in Figure \ref{fig:computational_cost}, increasing the sparse grid level improves cumulative rewards, indicating more efficient decision-making when we have better value function estimates. However, lower sparse grid levels (i.e. $q=1,2$) were insufficient to capture the tractable state-space uncertainties and failed to solve the problem. Moreover, the planning computational time increased proportionally with the sparse grid level while cumulative rewards quickly plateaued. 

We also explored using RBPF in POMCPOW by randomly sampling tractable states from the Gaussian distributions attached to each particle. In each tree iteration, this approach underperformed compared to SIRPF, indicating that random sampling from the marginalized distributions does not adequately capture the uncertainties in the tractable states when samples are insufficient. However, as the number of tree iterations increased, the performance of RBPF in POMCPOW improved and reached a comparable level of performance as SIRPF. Notably, POMCPOW, using either SIRPF or RBPF, consistently performed worse than RB-POMCPOW under the same computational time. Even at 1000 tree iterations, its performance was still as good as that of RB-POMCPOW with a sparse grid level of 3 and 50 tree iterations, which was shown to be approximately seven times faster in planning.

\begin{table}[!t]
    \caption{Belief Update Computational Time Per Step}
    \centering{}
    \begin{tabular}{lcc}
        \toprule
        \textbf{Particle Count} & \textbf{RBPF (ms)} & \textbf{SIRPF (ms)} \\
        \midrule
        100 & 9.073  & 4.336  \\
        1,000  & 90.92 & 32.14 \\ 
        10,000 & 910.2   & 315.4  \\
        \bottomrule
        \vspace{0.5mm}
    \end{tabular}
    \\
     \raggedright
    \footnotesize
     \textit{Note:} RBPF shows higher computational costs compared to SIRPF due to the added complexity of analytical filters. The time required for belief updates increases linearly with the number of particles.
      \vspace{-0.4cm}
     \label{tab:belief_update}
\end{table}

\section{Discussion}

The higher $N_{ESS}$ observed in RBPF, even with fewer particles, demonstrates its efficiency in state estimation within the POMDP model. In contrast, SIRPF required a substantially larger number of particles to estimate all three states and still exhibited low $N_{ESS}$. Additionally, we had to employ adaptive resampling and regularization by introducing a small amount of noise after the resampling step to prevent particle degeneracy. This highlights common issues in particle filtering and illustrates how RBPF can be more efficient and effective in addressing these limitations.

Although RBPF incurs higher computational costs for belief updates due to the added complexity of analytical updates, these costs are mitigated by reducing the number of particles. Also, our results show that RB-POMCPOW, using a sparse grid level of $3$ with $50$ tree iterations, achieved comparable performance to POMCPOW using 1000 tree iterations and was approximately seven times faster. Moreover, RB-POMCPOW outperformed POMCPOW under the same computational time at each sparse grid level. This validates the computational advantages of RBPF in both planning and belief updates without sacrificing accuracy or cumulative rewards. 

Future research could explore different Askey family distributions for sparse grid quadrature or simpler analytical updates, such as the Extended Kalman Filter (EKF), to further reduce computational time and apply the RB-POMDP framework to more complex problems. Since the number of particles needed for SIRPF to avoid particle collapse can grow “super-exponentially” with the state dimension \cite{c3}, this work presents a compelling alternative for use in those complex models.

Lastly, while this generic framework gives flexibility to practitioners, it also introduces additional parameters that need to be carefully considered and calibrated. This requires significant time and effort to fine-tune the model. One promising way to mitigate this burden is using the Bayesian optimization strategies proposed in \cite{c11} that could be employed to automate the tuning of Kalman filters. By adopting such methods when using Kalman filters for analytical updates, we could develop a more efficient framework with reduced reliance on heuristic calibration. 

\section{Conclusion}

In this study, we introduced the RB-POMDP framework that leverages RBPFs to enhance belief estimation and decision-making processes. Our flexible approach enables practitioners to incorporate analytical methods tailored to the distribution of tractable states and provides a generic framework that can be used for various applications. Additionally, the presented RB-POMCPOW planning algorithm allows users to balance accuracy and computational efficiency by selecting appropriate integration methods. The effectiveness of this approach was demonstrated in a localization problem where we employed UKF with Gaussian Hermite quadrature and Smolyak sparse grid. We believe this work highlights the potential of the RB-POMDP framework and opens the door to more computationally viable solutions in complex decision-making scenarios.

\appendix
\section{Appendix}
\label{appendix}
\subsection{RB-POMCP}
The following pseudocode outlines the RB-POMCP algorithm. The code is adapted from \cite{c4} with the modified code written in blue. The \textsc{Rollout} function is the same as the one defined in Algorithm \ref{alg:RB-POMCPOW}.
\begin{algorithm}
\caption{RB-POMCP}
\begin{algorithmic}[1]
\STATE \textbf{procedure} \textsc{Simulate}($p$,$h$, $depth$)
\STATE \quad \textbf{if} $\gamma^{depth} < \epsilon$ \textbf{then}
\STATE \quad \quad \textbf{return} 0
\STATE \quad \textbf{end if}
\STATE \quad \textbf{if} $hao \notin T$ \textbf{then}
\STATE \quad \quad \textbf{for all} $a \in A$ \textbf{do}
\STATE \quad \quad \quad $T(ha) \leftarrow (N_{\text{init}}(ha), V_{\text{init}}(ha), \emptyset)$
\STATE \quad \quad \textbf{end for}
\STATE \quad \quad \textcolor{blue}{\textbf{return} \textsc{Rollout}$(p, h, depth)$}
\STATE \quad \textbf{end if}
\STATE \quad $a \leftarrow \arg\max_{b} \left[ V(hb) + c\sqrt{\frac{\log N(h)}{N(hb)}} \right]$
\STATE \quad \textcolor{blue}{$(\hat{s'}, \hat{o}, \hat{r}) \leftarrow \mathbb{E}[G(p, a)]$} \hfill \textcolor{black}{$\triangleright$ Quadrature Techniques}
\STATE \quad \textcolor{blue}{$p' \leftarrow$ \textsc{AnalyticalUpdate}$(p, \hat{s'}, \hat{o}, a)$}
\STATE \quad \textcolor{blue}{$R \leftarrow \hat{r} + \gamma$ * \textsc{Simulate}($p'$, $ha\hat{o}$, $depth + 1$)}
\STATE \quad $N(h) \leftarrow N(h) + 1$
\STATE \quad $N(ha) \leftarrow N(ha) + 1$
\STATE \quad $V(ha) \leftarrow V(ha) + \frac{R - V(ha)}{N(ha)}$
\STATE \quad \textbf{return} $R$
\STATE \textbf{end procedure}
\end{algorithmic}
\end{algorithm}

\end{document}